\documentclass[twocolumn]{article} 
\usepackage{times}
\usepackage{soul}
\usepackage{url}
\usepackage[hidelinks]{hyperref}
\usepackage[utf8]{inputenc}
\usepackage[small]{caption}
\usepackage{graphicx}
\usepackage{amsmath}
\usepackage{amsthm}
\usepackage{booktabs}
\usepackage{algorithm}
\usepackage[switch]{lineno}

\usepackage{listings}

\usepackage{longtable}
\usepackage{algorithm}
\usepackage{algpseudocode}

\usepackage{natbib}
\usepackage{authblk}

\title{Generating Grounded Responses to Counter Misinformation via Learning Efficient Fine-Grained Critiques}

\author[1]{Xiaofei Xu}
\author[1]{Xiuzhen Zhang\thanks{Corresponding author.}}
\author[1]{Ke Deng}

\affil[1]{School of Computing Technologies, RMIT University, Melbourne, Australia}
\affil[ ]{\texttt{xiaofei.xu@ieee.org}, \texttt{\{xiuzhen.zhang, ke.deng\}@rmit.edu.au}}

\date{}

\begin{document}

\maketitle
\begin{abstract}
Fake news and misinformation poses a significant threat to society, making efficient mitigation essential. 
However, manual fact-checking is costly and lacks scalability. 
Large Language Models (LLMs) offer promise in automating counter-response generation to mitigate misinformation, but a critical challenge lies in their tendency to hallucinate non-factual information. 
Existing models mainly rely on LLM self-feedback to reduce hallucination, but this approach is computationally expensive. 
In this paper, we propose MisMitiFact, \underline{Mis}information \underline{Miti}gation grounded in \underline{Fact}s, an efficient framework for generating fact-grounded counter-responses at scale. 
MisMitiFact generates simple critique feedback to refine LLM outputs, ensuring responses are grounded in evidence. 
We develop lightweight, fine-grained critique models trained on 
data sourced from readily available fact-checking sites to identify and correct errors in key elements such as numerals, entities, and topics in LLM generations. 
Experiments show that MisMitiFact generates counter-responses of comparable quality to LLMs' self-feedback while using significantly smaller critique models. 
Importantly, it achieves $\sim$5x increase in feedback generation throughput, making it highly suitable for cost-effective, large-scale misinformation mitigation. 
%Our code and human evaluation results are available at~\url{https://github.com/xxfwin/MisMitiFact}. 
Code and LLM prompt templates are at \url{ https://github.com/xxfwin/MisMitiFact}. 

% \url{https://anonymous.4open.science/r/MisMitiFact-97DF9999}.\todo{update link} 
\end{abstract}

\section{Introduction}\label{sec:introduction}
% Misinformation can lead to misunderstandings, misconceptions, and even violence. 
% Research has shown that misinformation spreads faster and farther than truthful information~\cite{vosoughi2018spread}. 
% Professional fact-checkers and journalists can fact-check popular claims and publish their verdicts about the veracity of these claims. However, manual fact-checking requires labourious efforts by experts and is time-consuming, and thus typically focuses on only popular false claims. 
Misinformation spreads faster and farther than truthful information \cite{vosoughi2018spread}, posing significant risks to public health, trust, and society. While professional fact-checkers and journalists provide reliable veracity assessments, their efforts are labor-intensive and often focus only on popular claims. 
Automated fact-checking~\cite{guo2022survey,wang2023explainable,zeng2024justilm} have been proposed to identify misinformation at scale, but their focus has primarily been on veracity prediction rather than generating direct counter-responses in real-time. 
The widespread propagation of misinformation, especially on social media platforms, calls for the automated generation of factually grounded counter-responses in real-time and at scale.  
% Despite these efforts, research shows that various ground-zero misinformation still spreads wildly, especially on social media platforms. It is therefore critical to develop approaches to automatically generate responses to counteract misinformation in real-time and at scale. 

The advancement of Large Language models (LLMs) offers the opportunity for automated generation of counter-responses at scale. 
% Existing studies on the automated generation of counter-responses to false claims have been reported in the literature~\cite{he2023reinforcement} but they have a critical issue -- the factuality of the generated content is largely overlooked. Targeting counter-responses to engage users in social media conversations, He et al.~\cite{he2023reinforcement} proposed to employ LLM for counter-response generation, but their focus is on the language quality -- polite, containing evidence, of refutation attitude, and retaining text fluency and relevancy. Consequently, the generated counter-responses may contain non-factual information. 
%Existing studies on the automated generation of counter-responses to misinformation, such as the work by 
Leveraging LLMs, 
\cite{he2023reinforcement} proposed to employ LLMs for counter-response generation, but their focus is on the language quality -- relevant, fluent, polite and of refutation attitude. 
However, the critical issue of information factuality, or hallucination, remains largely overlooked. 
%hallucination -— the generation of non-factual or incorrect information. 
More broadly, studies on the mitigation of LLM hallucination are reported in the literature. 
Retrieval-Augmented Generation (RAG) approaches augment the input context for LLM generation with retrieved knowledge~\cite{lewis2020retrieval,yu2023improving,shi2023replug} to confine LLM generation and reduce hallucination. 
Originating from Chain of Thought (CoT) reasoning~\cite{kojima2022large,zhang2023language}, various prompting strategies are proposed to enhance LLM reasoning capabilities to reduce hallucination. 
Recently, LLM self-feedback strategy~\cite{madaan2024self,akyurek2023rl4f} is proposed to refine LLM generation, demonstrating strong performance to reduce hallucination. 
None of these studies directly address the issue of information factuality in LLM outputs. 
Importantly, all these approaches require additional LLM runs to generate reasoning steps or feedback, which is computationally expensive. 

%In contrast, Factual Error Correction (FEC) aims to identify and correct inaccuracies directly with post-editing correctors or retrieval-based systems~\cite{thorne2021evidence,lee2022factual}. While these approaches are effective for correcting factual errors for abstractive summarization, they do not consider the refutation of the claim, which is a key aspect of our problem. 

\begin{figure*}[tb]
\begin{center}
 \includegraphics[width=0.95\linewidth]{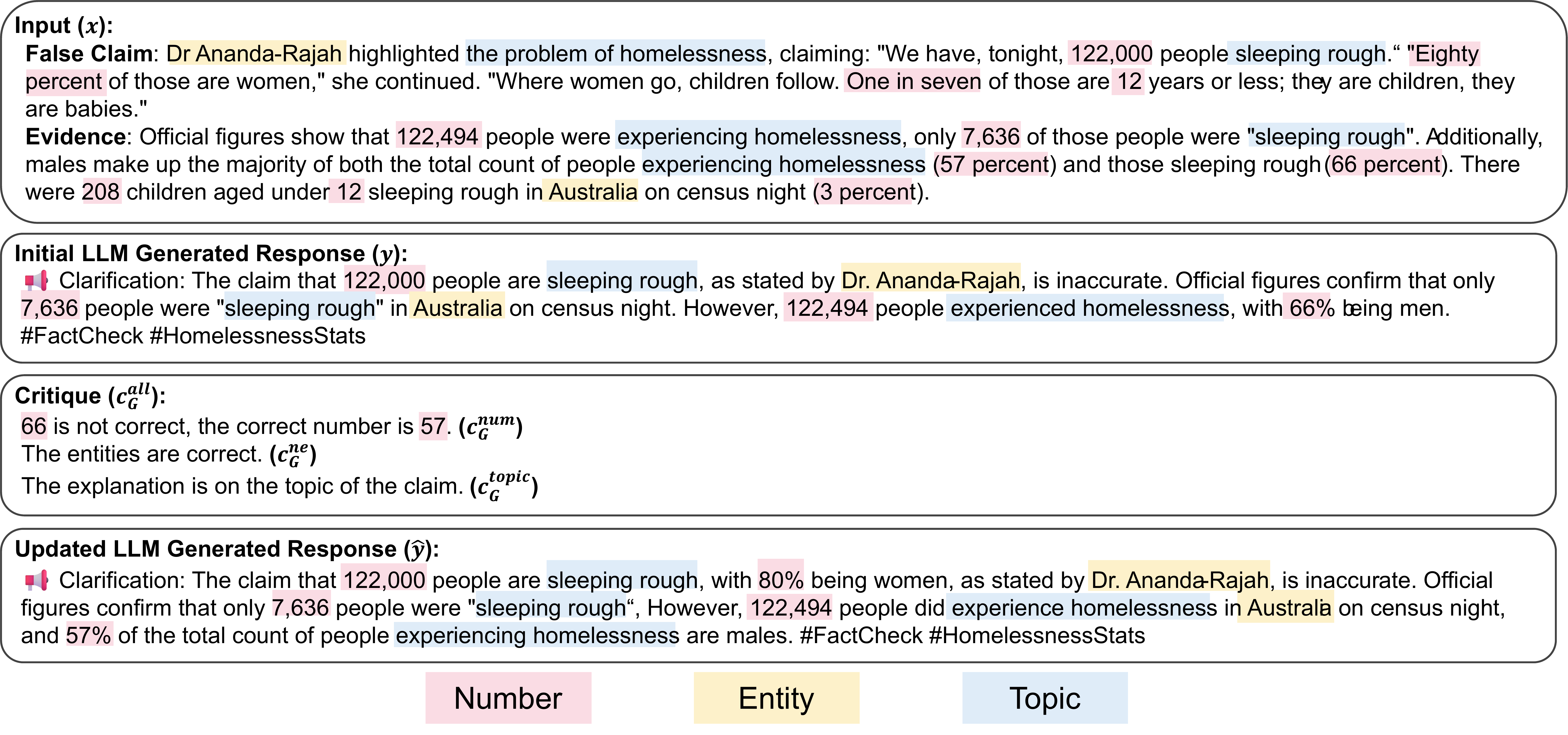} 
\caption{An example for MisMitiFact counter-response generation. 
Claim and evidence as input, 
initial LLM generation, 
the element-based critique feedback prompt on the initial generation, and updated LLM generation.
%from the critique prompt.   
Colours indicate different types of elements.}\label{fig:critique_example}
\end{center}
\end{figure*}

% ***** INFORMATION CORRECTNESS ****

% Generally hallucination, or incorrect information, in LLM generation is widely recognised in the literature~\cite{ji2023survey}. 
% Studies on mitigating hallucination in LLM generation focus on either augmenting the input context for LLM generation with information retrieved from a knowledge base~\cite{lewis2020retrieval,es2023ragas,yu2023improving,shi2023replug} or directly enhancing the reasoning capability for LLM to generate correct information~\cite{kojima2022large,zhang2023language,wang2024t,madaan2024self,akyurek2023rl4f}. 
% Studies in the latter group include 
% Chain of Thought (CoT) prompting strategies, which instruct the LLM to generate intermediate reasoning steps to improve the reasoning ability for correct generations~\cite{kojima2022large,zhang2023language,wang2024t}, 
% or feedback-based prompting strategies, which provide LLMs with prompts of feedback either from the LLM itself~\cite{madaan2024self} or from another feedback model~\cite{akyurek2023rl4f,lee2023rlaif} so that the LLM iteratively revises its output for better generation. 
% However, none of these studies directly address the issue of information correctness in the generated content. In contrast, Factual Error Correction (FEC) aims to identify and correct inaccuracies directly with post-editing correctors or retrieval-based systems~\cite{cao2020factual,lee2022factual}. While these approaches are effective for correcting factual errors for abstractive summarization, they do not consider the refutation of the claim, which is a key aspect of our problem. 

Recent research ~\cite{barron2020overview} shows that fact-check-worthy claims typically fall into two categories: potentially harmful claims attacking individuals, organizations, nations, or social groups and time-sensitive claims concerning prominent people, organizations, countries and events. 
Fact-checking such claims requires precise comparisons of key elements -- such as numbers, entities, and topics -- against factual evidence. For instance, as shown in Figure~\ref{fig:critique_example}, a false claim contains key elements such as numbers (e.g. 122,000, Eighty percent), entities (e.g. Australia) and topics (e.g. the problem of homelessness) that must be verified. 
Here an element refers to an individual piece of information that can be assessed for qualities such as factuality, correctness or relevance. 
%Our analysis of the PubHealth and COVID-19 Vaccine datasets reveals that false claims and their corresponding counter-responses frequently contain such elements (Table~\ref{table:datastat} in the Experiments section).

% Indeed, on both the Pubhealth and COVID-19 Vaccine datasets in our experiments, we observe a high proportion of number and entity elements in false claims and the corresponding counter-responses (Table~\ref{table:datastat} in the Experiments section). For example in the popular Pubhealth dataset about biomedical subjects and infectious disease, each false claim on average contains 0.51 number elements and 0.97 entity elements respectively. Correspondingly each ground truth response contains 0.62 number elements and 1.26 entity elements respectively. 

In this paper, we propose \textit{MisMitiFact} -- \underline{Mis}information \underline{Miti}gation grounded in \underline{Fact}s, an efficient framework for generating fact-grounded counter-responses to misinformation at scale. Unlike existing approaches that rely on expensive LLM inference for general feedback, our method introduces lightweight, fine-grained critique models that directly pinpoint factual errors about elements such as numbers, named entities, and topics. 
In our framework, critique models for factual errors about specific elements in the initial generation are trained using data within the readily available fact-checking articles from fact-checking sites. 
% Factual description and the related evidence from those fact-checking articles are used to build training data for the critique models, not requiring human-annotated critiques or counter-responses. 
Figure~\ref{fig:critique_example} shows how MisMitiFact works. The input $x$ consists of the misinformation claim and the evidence. We collect an initial counter-response, $y$, from the LLM, then obtain feedback from three critique models, $c_G^{all} = \{c_G^{num}, c_G^{ne}, c_G^{topic}\}$, which respectively evaluate the counter-response on numbers ($c_G^{num}$), entities ($c_G^{ne}$), and topics ($c_G^{topic}$). The LLM then takes the initial counter-response and the critiques into consideration and generates a refined counter-response, $\hat{y}$. 
%xf31: DONE
%jz31: in the figure, Intitial generated response > Intinitial LLM generated reponse; Updated generated reponse > updated LLM generated response. 
%Note that our critique models about numbers, entities and topics are trained to point out errors in the initial generated response ($y$). If no error is found in $y$, the critique model will output a pre-defined text indicating no errors. 

% By focusing on element-wise corrections and leveraging automatically generated training data, MisMitiFact achieves comparable performance to LLM-based baselines while significantly reducing computational and human effort.  

% In this paper, we propose to employ LLMs to generate counter responses grounded in evidence to debunk false claims, and we specifically address the hallucination issue. 
% Research~\cite{chen2021improving,ji2023survey} also shows that many LLM generated contents contain incorrect information about elements such as numbers, named entities and topics. 
% Different from existing prompting strategies to enhance generation in general~\cite{akyurek2023rl4f}, we propose a fine-grained critique feedback prompting strategy to directly pinpoint factual errors about numbers, topics and entities to enhance the information correctness and factuality for LLM generation. 

% Our system \textit{MisMitiFact} -- \underline{Mis}information \underline{Miti}gation grounded in \underline{Fact}s can automatically generate counter-responses grounded in facts from the given evidence to counter false claims. 

%Our contributions are: 
To our best knowledge, we are the first to introduce a task for counter-response generation grounded in factual evidence. 
We further propose a simple critique feedback approach for LLMs to refine generation and ensure information factuality. 
%fine-grained feedback critiquing errors in number, entity and topic elements. 
We train lightweight, efficient critique models targeting element-wise information correctness, the models are trained on data sourced from readily available fact-checking sites and do not require human-annotated critiques or counter-responses. 
Experiments on two real-world datasets show that our system MisMitFact can generate grounded counter-responses of quality comparable to the state-of-the-art LLM self-feedback approach while using significantly smaller critique models. 
Additionally, it achieves $\sim$5x increase in critique generation throughput, making our system highly suitable for cost-effective, large-scale misinformation mitigation.

\section{Related Work}\label{sec:literature}
Related work comes from three lines of research.
%misinformation mitigation, hallucination mitigation for LLMs and factual error correction. 

\subsection{Misinformation Mitigation}
A line of research on misinformation mitigation focuses on limiting the dissemination of false information while promoting corrective or clarifying truthful information~\cite{farajtabar2017fake,saxena2020mitigating,goindani2020social,xiaofei2022,xu2024harnessing}. These studies typically assume that counter responses containing truthful information are given.
%Their focus is not on the mitigation of contents. 
%
%The impressive ability of LLMs for content generation presents novel opportunities to generate content to mitigate misinformation. 
Another line of research focuses on automated fact-checking -- predicting the veracity of claims based on given claims and evidence~\cite{guo2022survey,wang2023explainable,zeng2024justilm}. 
%or generate personalized news content tailored to counteract misinformation~\cite{he2023reinforcement}. 
% In automated fact-checking, ~\cite{leippold2024automated} demonstrate the potential of leveraging large language models to fact-check the climate change claims by building a Mediator-Advocate system.
%\cite{wang2023explainable} and \cite{zeng2024justilm} proposes to build explainable automated fact-checking systems that leverage external knowledge and LLM reasoning capabilities. 
% explainable automated fact-checking systems by introducing external knowledge or document database and leveraging the reasoning ability of large language models to generate explainable fact-checking of claims. 
However, their focus is not on generating counter-responses to directly refute false claims. 
%However, these approaches often overlook the critical issue of ensuring the factual correctness of the generated content.

Some recent studies~\cite{he2023reinforcement} focus on counter-response generation. Their system MisinfoCorrect employs LLMs to generate counter-responses to COVID-19 vaccine misinformation. While their work models politeness, refutation attitude, and fluency within responses, it does not address the factual correctness of counter-responses, which is a critical issue and the focus of our research. 
%for using LLM-generated content for misinformation mitigation. 
% Targeting for a personalized counter-misinformation message generation, a reinforcement learning-based framework called MisinfoCorrect for generating counter-responses to COVID-19 vaccine misinformation on social networks has been proposed~\cite{he2023reinforcement}. However, they only model politeness, factuality, refutation attitude, and text fluency as rewards and utilize reinforcement learning to fine-tune the generation model. Their work has not considered the evidence for counter-response generation and does not focus on the factual correctness of the generated counter-response, which is a critical issue for using LLM-generated content for misinformation mitigation. 

\subsection{Mitigation of LLM Hallucination}
% It is widely recognized that large language models are susceptible to occasional inaccuracies or hallucinations~\cite{ji2023survey}, where they generate content that is not entirely rooted in evidence~\cite{akyurek2023rl4f}. 
%Most existing studies on LLM hallucination mitigation 
A line of research focuses on improving the generation process and broadly includes two classes of studies. 
% ~\cite{maynez2020faithfulness,ji2023survey}
One class is based on RAG~\cite{lewis2020retrieval,yu2023improving,shi2023replug}, which aims to reduce LLM hallucinations by augmenting the LLM input context with retrieved external knowledge base. 
Another class leverages CoT prompting~\cite{kojima2022large,zhang2023language}, which improves LLM reasoning capabilities to reduce hallucinations. 
While these methods enhance the overall quality of LLM generation, they do not directly address the information factuality. 

% which prompts the large language models to generate reasoning steps for the problem, LLM reasoning capability is improved to reduce hallucination in the generated content. Still, none of these two classes of studies directly address the correctness of information in the generated contents. 

Another line of research provides feedback to refine LLM generation and improve its quality~\cite{schick2022peer,akyurek2023rl4f,lee2023rlaif,yu2023improving,madaan2024self}.
Early research using textual critique feedback prompts to improve LLM generation, where a separate critique model for generating critiques~\cite{schick2022peer,akyurek2023rl4f} is trained from human-written critiques as training data. 
%The critique model is trained via supervised fine-tuning or reinforcement learning, using human-written critiques as training data. 
A recent study called SELF-REFINE~\cite{madaan2024self} repeatedly calls an LLM to generate feedback on its own generation. 
%All these existing studies, however, focus on general feedback and do not specifically target factual errors in the generated content. 
%Moreover, they rely on expensive LLM inference to generate feedback. 
%which limits their scalability for large-scale misinformation counteraction. 
% however, use feedback to improve the overall quality of LLM generation and do not focus on correcting specific errors in the information in the generated contents. 
This approach not only generates general feedback but also incurs repeated LLM calls, which are computationally expensive. 
In contrast, our approach introduces lightweight, fine-grained critique models that generate simple critique prompts targeting factual errors in key elements (e.g., numbers, entities, and topics). Our lightweight critique models generate simple, element-wise critiques that incur significantly less computation cost and our MisMitiFact achieves comparable textual quality to LLM-based self-feedback. 
%while significantly reducing computational costs, making it more suitable for large-scale applications. 
% In contrast, our MisMitiFact is a prompt-based approach that utilizes critique prompts to mitigate hallucination in counter-response generation. Our critique models are focused on providing critiques for factual elements in LLM generation and experiments also show that our approach is better than general feedback for counter-response generation. 

\subsection{Factual Error Correction}
Our general idea of identifying and correcting errors in texts is somewhat related to the NLP task of Factual Error Correction (FEC) for abstractive text summarization~\cite{thorne2021evidence,lee2022factual}. 
FEC focuses on checking the factual correctness of summaries given source documents, a fundamentally different and much simpler problem setting that does not involve claims and their counter-responses. 

\begin{figure*}[tbh]
\begin{center}
 \includegraphics[width=0.95\linewidth]{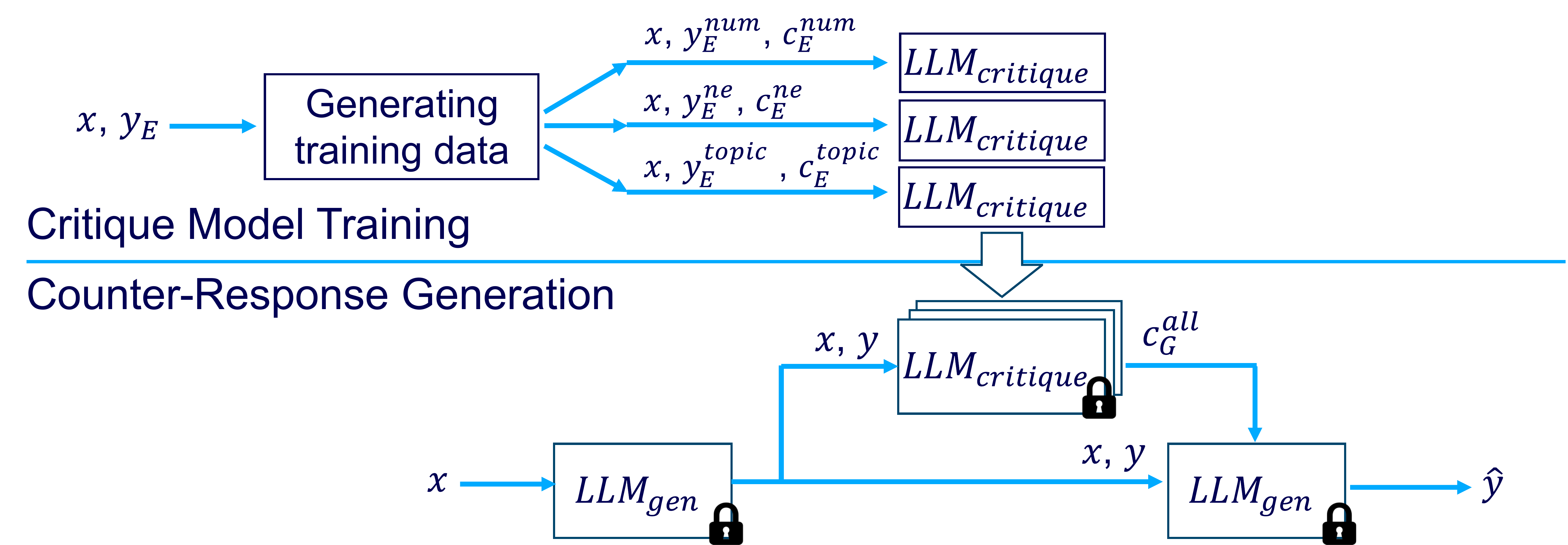} 
 % \includesvg[inkscapelatex=false, width=0.95\linewidth, keepaspectratio]{fig3.svg}
 \caption{The MisMitiFact framework has two phases: critique model training and counter-response generation. In the critique model training phase, 
 training data are automatically generated to train three $LLM_{critique}$ models on numbers, named entities, and topics. During the counter-response generation phase, a frozen LLM, $LLM_{gen}$, generates an initial counter-response, which is critiqued by the critique models $LLM_{critique}$, and the critiques are further used as feedback to prompt and refine $LLM_{gen}$ to generate the final counter-responses.}\label{fig:mismitifact}
\end{center}
\end{figure*}

\section{Problem Definition}\label{subsec:problemdef}

Our task is to generate counter-response text $y$ to debunk a misinformation claim $x_c$, given factual evidence text containing facts related to the claim, $x_e$. 
Counter-response $y$ is desired to have the following properties:
\begin{itemize}
	\item \textit{Faithful to the evidence}. Denote all statements in counter-response $y$ as $S = \{s_1, \cdots, s_N\}$ and all verifiable statements in $y$ as $V = \{v_1, \cdots, v_N\}$. The response $y$ is faithful to the evidence iff $\forall s_i \in y, s_i \mathrel|\joinrel= x_e$ and $S-V=\emptyset$. For a response to hold this property, all the statements in the response must be verifiable and be verified true ($\mathrel|\joinrel=$) according to $x_e$.
	\item \textit{Refute the misinformation}. For a successful refutation to the misinformation, the counter-response should satisfy the property of \textit{faithful to the evidence} and also satisfy the following property: Define all false statements in the misinformation claim $x_c$ as $F = \{f_1, \cdots, f_N\}$. Response $y$ refutes the misinformation iff $\forall f_i \in x_c, f_i \mathrel|\joinrel\neq y$. For a response to hold this property, all the false statements in the claim must be clarified as false ($\mathrel|\joinrel\neq$) by $y$. 
\end{itemize}

For counter-responses failing to satisfy the property of \textit{faithful to the evidence}, the response may contain non-factual information and is not acceptable when debunking misinformation. For counter-responses failing to satisfy the property of \textit{refute the misinformation}, the responses will not clarify the false claim and thus never counter the misinformation. 
%Targeting a successful debunking of misinformation, 
We aim for a framework that leverages LLMs to automatically generate counter-responses to refute false claims while correcting factual errors contradicting facts in evidence. 

% As most state-of-the-art LLMs do not have publicly available parameter weights, we assume frozen LLMs. 

% \begin{figure*}[tb]
% \begin{center}
%  \includesvg[inkscapelatex=false, width=0.85\linewidth, keepaspectratio]{fig4.svg}
% \caption{A quadrant diagram showing the counter-responses generated with different combination of properties. The counter-responses are generated based on the example shown in Figure \ref{fig:critique_example}. A counter-response with both \textit{faithful to the evidence} and \textit{refute the misinformation} is desired. }\label{fig:response_quadrant}
% \end{center}
% \end{figure*}

%In this section, we describe our approach, MisMitiFact, a misinformation counter-response generation approach that is grounded in the provided evidence. 
\section{Methodology}\label{sec:method}
We propose a prompt-learning approach to counter-response generation, focusing on critique feedback to prompt LLMs to correct errors within the initial generation. 
% and we are specifically interested in providing critique feedback to correct errors for LLM generation. 
Inspired by observations that fact-check-worthy claims often contain false information about elements such as numbers, named entities and topics, we propose to train small critique models to pinpoint errors and generate critique feedback for elements (See Figure~\ref{fig:critique_example} for an example). 
We also aim not to incur extra human annotations to train the critique models.  
%the automatic generation of training data for these critique models from readily available fact-checking articles at fact-check sites, removing the need for human-written critiques or counter-responses. 

\subsection{MisMitiFact}
Figure \ref{fig:mismitifact} shows our MisMitiFact framework, 
which can be summarized as follows: 
(1) A large language model with frozen weights, $LLM_{gen}$, is used to generate an initial counter-response to a misinformation claim. 
(2) The initial output of $LLM_{gen}$ is critiqued by a set of three trained small critique models, $LLM_{critique}$. (3) The text output of the critics, along with the initial input to $LLM_{gen}$, are used to prompt $LLM_{gen}$ to generate a refined output. 
In contrast to the existing critique-style feedback approach of~\cite{akyurek2023rl4f}, which relies on expensive human-written critiques, our approach trains lightweight, critique models on specific elements like numbers, named entities and topics. These models are trained using training data automatically generated from readily available fact-checking articles (Details in Section~4.2). 

% the proposed approach avoids training a single critique model from human-written critiques, instead training several critique models on specific elements like numbers, named entities and topics, using training data generated from the factual description of related evidence in the readily available fact-checking articles. 

MisMitiFact consists of two phases: critique model training and counter-response generation. 
In the critique model training phase, given input $x$ (consists of claim $x_c$ and evidence $x_e$) and explanation $y_E$, we generate training data comprising two types of instances to train $LLM_{critique}$: 
The \textbf{factual instances}, where the input $x$ and the original explanation $y_E$ is paired with 
%an affirmative indicator (e.g., ``The entities are correct") to train the model to confirm factual accuracy. 
a positive indicator like ``The entities are correct" to train the model to confirm factual accuracy.
% when no inaccuracies exist. 
The \textbf{counter-factual instances}, 
%where $y_E$ is altered to generate three types of counterfactual explanations 
%- numerically replaced $y_E^{num}$, named-entity replaced $y_E^{ne}$, and off-topic $y_E^{topic}$ — 
where the factual elements number, name entity and topic in $y_E$ are replaced with false number, false named entity and off-topic, generating 
$y_E^{num}$, $y_E^{ne}$, and $y_E^{topic}$.  
Each counter-factual instance is paired with input $x$ and critiques based on templates ($c_E^{num}$, $c_E^{ne}$, and $c_E^{topic}$), indicating the introduced factual errors. 
% In the critique model training phase, given input $x$ and explanation $y_E$, we generate three corresponding data points that are used to train $LLM_{critique}$: We modify numbers in $y_E$, yielding a false description $y_E^{num}$, and generate a corresponding critique $c_E^{num}$ from a pre-defined template. Similarly, we create false description $y_E^{ne}$ and $y_E^{topic}$ by replacing named entities and inserting off-topic data respectively, again generating corresponding critiques, $c_E^{ne}$ and ${c_E^{topic}}$ via templates. 

% After preparing the training data, we train three corresponding $LLM_{critique}$ models. 
This generated structured training data ensures the $LLM_{critique}$ models learn both to validate factually explanations and to detect counterfactual explanations using critiques. 
For example, with a false claim $x_c = $\textit{``122,000 people sleeping rough."} and collected evidence $x_e = $\textit{``Official figures show that 122,494 people were experiencing homelessness, only 7,636 of those people were sleeping rough."}, a journalist might write an explanation $y_E$ as \textit{``only 7,636 people were sleeping rough."}.  
We will replace the number \textit{7,636} in $y_E$ with other numbers provided in evidence $x_e$, that is \textit{122,494}. Thus, the $y_E^{num}$ will be \textit{``only 122,494 people were sleeping rough."} with a corresponding $c_E^{num}$ as \textit{``122,494 is not correct, the correct number is 7,636"}. The same idea also applies to $(y_E^{ne}, c_E^{ne})$ and $(y_E^{topic}, c_E^{topic})$. 

% and the details can be found at Section~\ref{appendix:critiquedata}. 

%After the critique model training, we enter the response generation stage. In this stage, 
At the inference stage, 
all models, including the trained critique model $LLM_{critique}$ and the generation model $LLM_{gen}$, are frozen. When generating the critique for the initially generated counter-response, the critique is given by all the critique models in $LLM_{critique}$ as $c_G^{num}, c_G^{ne}$ and $c_G^{topic}$ respectively. We concatenate all the critiques generated by $LLM_{critique}$ as $c_G^{all}$ and feed into the prompt template of $LLM_{gen}$. Prompt templates of $LLM_{gen}$ are at {\url{https://github.com/xxfwin/MisMitiFact}}.

% *** All the factual descriptions and counter-responses should be explanations
\subsection{Training Data Generation for Critique Models}
\label{appendix:critiquedata}
%To reduce the human effort in building the training dataset for the critique models in $LLM_{critique}$, we introduce several techniques 
We next describe how to automatically generate training data for critique models from readily available fact-checking articles. 
Fact-checking articles usually include four components: (1) a claim, (2) human-curated evidence gathered from credible sources to fact-check the claim, (3) a verdict  ("true", "false" or "mixed"), and (4) human explanation to support the fact-check verdict for the claim. 
%As we focus on the false claims, we only include the false claims. 
As we focus on the false claims, we only include the false claims. 

To generate training data for critique models, we focus on using the original claim $x_c$, the human-curated evidence $x_e$ and the explanation $y_E$ to generate two training components: 
a) \textit{factual instances}: We pair the explanation $y_E$ with its corresponding claim $x_c$ and evidence $x_e$, appending affirmative critiques (e.g., ``The numbers/entities are correct" or ``The explanation is on the topic of the claim"). These instances train the model to confirm factual accuracy when no factual errors exist. 
b) \textit{counter-factual instances}: The $y_E$ will be modified to generate three types of false explanations - numerically replaced $y_E^{num}$, named-entity replaced $y_E^{ne}$, and off-topic $y_E^{topic}$. 
For $y_E^{num}$, we systematically replace number $q^y$ in an explanation $y_E$ with a different number $q^x$ from the evidence text $x_e$ and then create critique $c^{num}_E$ following template "$q^x$ is not correct, the correct number is $q^y$". 
Similarly, for $y_E^{ne}$, we replace named entity $q^y$ in an explanation $y_E$ with a different named entity $q^x$ from the evidence text $x_e$ and then create critique $c^{ne}_E$ following template "$q^x$ is not correct, the correct text is $q^y$". 
For $y_E^{topic}$, we prompt Google Gemini~\cite{team2023gemini} to generate off-topic messages based on the given evidence. The instruction is to generate off-topic messages that are different from the explanation but still related to the evidence. The training data for the topic critique model is generated by prompting \texttt{gemini-pro} using the prompt in Table~\ref{tab:prompt1}. 
For each claim $x_c$, we generate a maximum of 20 instances of numerically replaced and named-entity replaced explanations and generate 3 instances of off-topic explanations. Here, we utilize spaCy~\cite{spacy2} to perform named entity recognition to extract all named entities or numbers.

\begin{table}[h]
\centering
\begin{tabular}{p{0.95\linewidth}}
% \toprule
% \textbf{Prompt for generating training data for topic critique model}\\
% \midrule
\toprule
Your task is to rewrite the explanation so that it is off-topic to the provided claim but still on the topic of the provided facts. The output of your response in a single plain json list, please return the rewritten explanation as ``rewritten\_explanation" and reason as ``reason" in the json list. Please also try to keep a similar length of the provided claim. \\
Here is the template of the reason: The claim is about \textless on topic part of claim\textgreater, but the explanation is not correct because it is about \textless off topic part of explanation\textgreater. \\
\\
This is the claim:\\
\{claim\}\\
\\
This is the explanation:\\
\{explanation\}\\
\\
This is the facts: \\
\{evidence\} 
\\
\bottomrule
\end{tabular}
\caption{Prompt for generating training data for topic critique model}
\label{tab:prompt1}
\end{table}

% Please provide 3 examples.

\begin{table*}[ht]
\begin{center}\small
\begin{tabular}{lccccccc}
\cline {1-7}
& \multicolumn{3}{c}{ PUBHEALTH } & & \multicolumn{3}{c}{ COVID-19 Vaccine } \\
\cline { 2 - 4 } \cline { 5 - 7 } 
Metric & Claim & Evidence & Explanation  & Claim & Evidence & Explanation \\
\cline {1-7}
Avg. No. of Tokens & 23.84 & 1146.69 & 32.27 & 53.20 & 1547.00 & 37.20 \\
Avg. No. of Numbers & 0.51 & 23.49 & 0.62 & 0.49 & 10.00 & 0.07 \\
Avg. No. of Entities & 0.97 & 36.04 & 1.19 & 1.26 & 9.00 & 0.60 \\
\cline {1-7}
\end{tabular}
\end{center}
\caption{Statistics of the PUBHEALTH dataset and COVID-19 Vaccine dataset.}
\label{table:datastat}
\end{table*}

\section{Experiments}\label{sec:experiments}
%We evaluate MisMitiFact against baselines from both the misinformation counter-response generation and hallucination mitigation literature. 
Experiments are conducted on a cluster where each node has 32 cores, 128G memory and is equipped with an NVIDIA Geforce RTX 3090. All deep neural networks are implemented using Transformers~\cite{wolf2019huggingface} under the support of PyTorch~\cite{paszke2019pytorch}. 

% including critique model selection and hyperparameter settings, see Appendix ~\ref{appendix:implementation}. 

%\noindent \textbf{Generation Models.}\label{generationmodels} 
\subsection{MisMitiFact and Baselines}

% For the implementation details, w 
For MisMitiFact implementation, we fine-tuned the \texttt{T5-large}~\cite{raffel2020exploring} for the critique models ($LLM_{citiqiue}$ in Fig.~\ref{fig:mismitifact}), 
with a learning rate of 1e-5, 5 epochs, temperature of 1.0 and output length of 5-30 tokens. The overall training time of the three critique models is around 30 hours. 
%the implementation of supervised fine-tuning of t5-large is modified from~\cite{akyurek2023rl4f}. 
%The supervised fine-tuning has a learning rate of 1e-5, a total number of epochs of 5. 
%Generated content is forced to have a minimal length of 5 and a maximal length of 30. 
%For all generation models in this work, we used the GPTQ version of the models. Since popular social media platforms usually have a limitation on the post length, to reflect that limit, the generation configuration is set with a maximum number of new tokens as 150 under a temperature of 1.0. 
Generate critiques are set to maximum 150 tokens. 
% Since the counter-response generation task involves natural language generation models (denoted $LLM_{gen}$ in our earlier discussion), we perform experiments on two popular instruction-tuned LLMs: 
% \begin{itemize}
% 	\item \emph{Vicuna}~\cite{zheng2024judging} is a popular open-source large language model developed by LMSYS. Vicuna is further fine-tuned on LLaMA-2 using extra datasets collected. 
%     % and supporting long context sizes of up to 16k. 
%  %A highlight of Vicuna version 1.5 is that is supports long context sizes of up to 16k, which is desirable in tasks like counter-response generation.
% 	\item \emph{LLaMA2}~\cite{touvron2023llama} is a popular open-source large language model developed by Meta. Here, we use the \textit{chat} version to support our instructions used in the prompt.
%     % that is fine-tuned on instruction data 
% \end{itemize}
For the generation model of MisMitiFact ($LLM_{gen}$ in Fig.~\ref{fig:mismitifact})  we experimented with two popular open-source LLMs Vicuna-1.5~\cite{zheng2024judging} and LLaMA-2~\cite{touvron2023llama}. 
%\begin{itemize}
% 	\item \emph{Vicuna}~\cite{zheng2024judging} is a popular open-source large language model developed by LMSYS. Vicuna is further fine-tuned on LLaMA-2 using extra datasets collected. 
%     % and supporting long context sizes of up to 16k. 
%  %A highlight of Vicuna version 1.5 is that is supports long context sizes of up to 16k, which is desirable in tasks like counter-response generation.
% 	\item \emph{LLaMA2}~\cite{touvron2023llama} is a popular open-source large language model developed by Meta. Here, we use the \textit{chat} version to support our instructions used in the prompt.
%     % that is fine-tuned on instruction data 
% %\end{itemize}

%\subsection{Baselines}
%\noindent \textbf{Baselines.}\label{baselines} 
We compare MisMitiFact against 5 baseline models. 
For a fair comparison, we also use the same Vicuna and LLaMA models as the generation models for all baselines. 
%from both the misinformation counter-response generation and feedback to improve LLM generation approaches. 
%Since there is limited research in misinformation counter-response generation, 
%we only tested one baseline in this research area:
\begin{itemize}
\item    
    \emph{MisinfoCorrect}~\cite{he2023reinforcement}  is a recent model for misinformation counter-response generation model.
    %based on GPT2 that directly generates counter-responses for misinformation claims. 
    %For fair comparison, we used \emph{Vicuna} and \emph{LLaMA2} as the base LLMs for our implementation of MisInfoCorrect. 
\item
    \emph{MisinfoCorrect w/ Evidence} is an extension of MisinfoCorrect, including evidence in the input context. 
    %Instead of using the MisinfoCorrect to refute the claim by only using the claim as the input. 
    % , we include the evidence to obtain better faithfulness of the generated response. The prompt template will be using the one to generate the initial response in MisMitiFact. 

    \item  
    \emph{SELF-REFINE}~\cite{madaan2024self} is a recent self-feedback approach %that leverages the ability of large language models to improve the quality of outputs through iterative feedback and refinement. It 
    that uses an LLM for generation and calls the LLM again to generate feedback and refine its generation. 
    %and has been proven effective in many tasks. 
    For a fair comparison, we include evidence in the input context.
    %and run in a zero-shot setting. 
% Due to the limited input length, we only apply SELF-REFINE in a zero-shot setting. 
% Write some sentences to be fair to their model, we put evidence as part of input. 
 
%As the generation model of this baseline has a limited ability, we take the idea of directly generating a counter-response based on the claim only and 
%\end{itemize}
%Two more baselines can be formed by applying existing hallucination mitigation techniques in large language models: 
%\begin{itemize}

    \item 
    \emph{Chain-of-Thought} (CoT) is a popular prompt technique that aims to reduce hallucination in LLMs. We applied zero-shot CoT~\cite{kojima2022large} to instruct the LLMs to solve the task step by step.

    \item 
    \emph{Plug-and-plug REtrieval FEEDback} (REFEED)~\cite{yu2023improving} is a RAG-based model,
%    uses a document retrieval technique to help reduce hallucination. 
where the retrieved documents are fed back to the initially generated content, allowing the generation model to refine the content. For our task, the evidence is used as the retrieved documents.
\end{itemize}

%It is natural to include directly supervised fine-tuning of the generation model as a baseline, however, this approach is not always applicable when we have limited access to the generation model. 

% Note that under our problem setting of a frozen LLM as the generation model, the straightforward approach of supervised fine-tuning with human-written gold standard counter-responses to train the generation model is not applicable. Nevertheless, for completeness, we implemented the fine-tuning approach for our task.  
% As shown in Appendix~\ref{appendix:sft}, the fine-tuning approach has poor performance, demonstrating the complexity of the counter-misinformation response generation task. 

% In our setting with a frozen LLM as the generation model, supervised fine-tuning with human-written counter-responses is not feasible. 
% Despite this, we implemented the fine-tuning approach for completeness, but it performed poorly (see Appendix~\ref{appendix:sft}), highlighting the complexity of the counter-misinformation response generation task. 
Note that we have not included as baseline models for critic-style feedback to improve LLM generation discussed in Section~\ref{sec:literature}, as they all require gold standard critiques, which are not available in our problem setting. %Similarly, we have not included factual error correction approaches in our baselines because they primarily focus on abstractive text summarization, which is different from our problem where the counter-response should refute the claim while remaining faithful to evidence. 

% or the factual error correction approaches
% For example, ~\cite{akyurek2023rl4f} proposed RL4F to utilize reinforcement learning to obtain a critique model for better generation performance. Due to a lack of gold standard critiques, we directly apply reinforcement learning following RL4F with the similarity between generated responses and gold standard responses as the reward.\todo{MD: This statement seems a bit loose. How can a text response be a reward? Do we mean that it's used as a preference?\\ XF: Added similarity between these two as a reward, trying to avoid mentioning we are using ROUGE score as reward} However, this results in the degeneration of the critique model with no improvement of the counter-response generation. 
% \cite{es2023ragas} proposed to use retrieval augmented generation to mitigate the hallucination of LLMs. However, this line of work is designed to train a retriever model and a generation model, which is outside the scope of this work. 

%\subsection{Data, Metrics and Annotation}
\subsection{Experiment Setup}

\noindent \textbf{Datasets:} Experiments were conducted on two datasets: 
\begin{itemize}
	\item PUBHEALTH~\cite{kotonya2020explainable} is a large fact-checking dataset containing 11.8K claims on public health topics. %Claims were collected from fact-checking sites and news review websites. 
    For each claim, there is evidence curated by journalists from credible sources, veracity labels (true, false, unproven, mixture) and counter-responses (called  ``explanations'') crafted by journalists. For our experiments, we use the 2153 claims with the ``false" or ``mixture" label. 
    %jz31: what's the difference between evidence and "fatual descriptions"?
    %xf31: maybe still call them counter-responses are more safe. This might also answer the question below. 
    % the discussion text as evidence, and explanations 
    % as factual descriptions of the evidence. 
    % as gold standard counter-responses. 
    % However, due to the various sources, some explanations in the PUBHEALTH dataset do not necessarily act as counter-responses but as factual descriptions of the evidence.  
	\item  COVID-19~\cite{he2023reinforcement} includes false claims about the COVID-19 vaccine from Twitter (now X) and gold standard counter-responses by crowdsourcing. For our experiments, we crawled the source claims based on tweet ID~\footnote{https://developer.twitter.com/en/docs/twitter-api} and got 307 false claims.
 %The dataset does not contain relevant discussions about the source claims from credible sources. 
 %Therefore, 
 To construct relevant evidence about the claims, we crawled the ``Facts about COVID-19 Vaccines" web page from the CDC website~\footnote{https://www.cdc.gov/coronavirus/2019-ncov/vaccines/facts.html}. The same evidence on the topic of the COVID-19 vaccine is used for all claims. 
\end{itemize}

%jz31: how ground-truth reponses are used for evaluation? responses = factual description earlier? 
We analysed the contents of the claim, evidence and explanation text to understand the characteristic differences between the two datasets. We found that the datasets share a similar count of numbers and entities in the claims, but the counts in the explanations differ a lot. 
%jz3: Below, Responses matter? These are ground truth responses? 
In COVID-19, the explanations contain nearly no numbers. This might be because the majority of false claims can be explained without quoting numbers. From the contents of the evidence, it can be seen that the evidence in PUBHEALTH contains more numbers and named entities, meaning that it is effectively more information-dense than the COVID-19 Vaccine dataset. Table~\ref{table:datastat} includes statistics of the two datasets used. 
For all models, we use 80\% of claims for training, 10\% for development and 10\% for testing. 
%\todo{Do we still need to mention this? Since we only evaluate 100 instances}

\begin{table*}[th]
\begin{center}\small
\renewcommand{\arraystretch}{0.8}
\begin{tabular}{lcccccc}
\toprule
Model  & Numerical$\uparrow$ & Entity$\uparrow$ & Faithfulness$\uparrow$ & Refutation$\uparrow$ & FActScore$\uparrow$ & Overall$\uparrow$\\ 
\midrule
\multicolumn{6}{c}{PUBHEALTH Dataset} \\
\midrule
\multicolumn{5}{l}{$LLM_{gen}$ = Vicuna} \\ 
\hspace{4mm} MisMitiFact (ours)     & \textbf{0.987}  & 0.873            & 0.881       & 0.716   &  0.733 & 0.838 \\
\hspace{4mm} MisinfoCorrect  & 0.924 & 0.741 & 0.661 & 0.550 & 0.540 & 0.683 \\
\hspace{4mm} MisinfoCorrect w/ Evidence  & 0.908  & \textbf{0.920}          & 0.890       & \textbf{0.777}  &  0.727 & \textbf{0.844} \\
\hspace{4mm} SELF-REFINE  & 0.822 & 0.861 & 0.835 & 0.652 & \textbf{0.752} & 0.784 \\
\hspace{4mm} CoT   & 0.673  & 0.888  & \textbf{0.892}  & 0.753  & 0.719  & 0.785 \\
\hspace{4mm} REFEED  & 0.931 & 0.762 &  0.668 &  0.621  & 0.619 & 0.720 \\
\midrule
\multicolumn{5}{l}{$LLM_{gen}$ = LLaMA2} \\ 
\hspace{4mm} MisMitiFact (ours)    & 0.889    & \textbf{0.871}          & \textbf{0.873}       & 0.711  & 0.705 & 0.810 \\
\hspace{4mm} MisinfoCorrect  & 0.742 & 0.672 & 0.606 & 0.510 & 0.495 & 0.605\\
\hspace{4mm} MisinfoCorrect w/ Evidence          & 0.889  & 0.862          & 0.854       & \textbf{0.781}  & 0.679 & 0.813 \\
\hspace{4mm} SELF-REFINE  & 0.917 & 0.854 & 0.871 & 0.751 & \textbf{0.744} & \textbf{0.827}\\
\hspace{4mm} CoT  &  \textbf{0.967} & 0.855 & 0.859 &  0.732  & 0.617 & 0.806 \\
\hspace{4mm} REFEED  & 0.733  & 0.729 & 0.674 &  0.574 & 0.484 & 0.639 \\
\midrule
\multicolumn{6}{c}{COVID-19 Vaccine Dataset} \\ 
\midrule
\multicolumn{5}{l}{$LLM_{gen}$ = Vicuna} \\ 
\hspace{4mm} MisMitiFact (ours)  & \textbf{0.987}  & \textbf{0.911}   & \textbf{0.840}  & 0.690 & 0.771 & \textbf{0.840} \\
\hspace{4mm} MisinfoCorrect    & 0.931 &  0.831 & 0.770 &  0.745 & 0.432 & 0.742 \\
\hspace{4mm} MisinfoCorrect w/ Evidence  & 0.983  & 0.862          & 0.814       & \textbf{0.763}  & 0.729 & 0.830 \\
\hspace{4mm} SELF-REFINE  & 0.975 & 0.880 & 0.789 & 0.678 & \textbf{0.777} & 0.828 \\
\hspace{4mm} CoT     & 0.928  &  0.859 & 0.820 &  0.745 & 0.665 & 0.803\\
\hspace{4mm} REFEED    & 0.756 & 0.783 & 0.690 & 0.705 & 0.352 & 0.657 \\
\midrule
\multicolumn{5}{l}{$LLM_{gen}$ = LLaMA2} \\ 
\hspace{4mm} MisMitiFact (ours)    & 0.933    & \textbf{0.869}          & \textbf{0.815}       & 0.717  & 0.686 & 0.804 \\
\hspace{4mm} MisinfoCorrect   & 0.816 & 0.731  & 0.630 & 0.703  & 0.363 & 0.649\\
\hspace{4mm} MisinfoCorrect w/ Evidence   & \textbf{0.964}  & 0.829          & 0.743       &  0.727  & 0.490 & 0.751\\
\hspace{4mm} SELF-REFINE  & 0.964 & 0.863 & 0.774 & \textbf{0.731}  & \textbf{0.697} & \textbf{0.806}\\
\hspace{4mm} CoT   &  0.619  & 0.809 & 0.727 & 0.685  & 0.450  & 0.658\\
\hspace{4mm} REFEED  & 0.890  & 0.769 & 0.698 & 0.683  & 0.399 & 0.688\\
\bottomrule
\end{tabular}
\end{center}
\caption{Experiment results on PUBHEALTH and COVID-19 using LLM-based metrics for factual correctness and refutation}
\label{table:mainresult}
\end{table*}

% \begin{table}[ht]
% \begin{center}
% \begin{tabular}{lccccccc}
% \cline {1-7}
% & \multicolumn{3}{c}{ PUBHEALTH } & & \multicolumn{3}{c}{ COVID-19 Vaccine } \\
% \cline { 2 - 4 } \cline { 5 - 7 } 
% Metric & Numerical & Named Entity & Topic & Numerical & Named Entity & Topic \\
% \cline {1-7}
% \multicolumn{7}{l}{$LLM_{gen}$ = Vicuna~\cite{zheng2024judging}} \\ 
% \hspace{4mm} Avg. No. in Total & 000 & 000 & 000 & 000 & 000 & 000 \\
% \hspace{4mm} Avg. No. of Mistakes & 000 & 000 & 000 & 000 & 000 & 000 \\
% \multicolumn{7}{l}{$LLM_{gen}$ = LLaMA2~\cite{touvron2023llama}} \\ 
% \hspace{4mm} Avg. No. in Total & 000 & 000 & 000 & 000 & 000 & 000 \\
% \hspace{4mm} Avg. No. of Mistakes & 000 & 000 & 000 & 000 & 000 & 000 \\
% \cline {1-7}
% \end{tabular}
% \end{center}
% \caption{Statistics of the mistakes in the initial response generated on the PUBHEALTH dataset and COVID-19 Vaccine dataset.}
% \label{table:initres_stats}
% \end{table}

% We further analysed the mistakes made by the large language models when generating the initial responses on both datasets. Table~\ref{table:initres_stats} includes statistics of the mistakes in the initial response generated by two different base large language models. We found that the count of mistakes in the initial responses on both datasets has a similar trend in terms of numerical and named entities. This reflects the observations in previous research~\cite{ji2023survey}. However, the results on the two datasets share a different count in terms of topic. This might be because of the wide topic range in the PUBHEALTH dataset. 

\noindent \textbf{Evaluation Metrics:}\label{evaluationmetrics} 
% The generated responses are evaluated by humans based on several desired properties. 
% In the real world, there is not one ground-truth counter-response for a given false claim, rather, there can be many -- each counter-response can counteract a different aspect of the claim -- and their semantics can be very different. 
% Therefore we focus on evaluating the factuality -- factual consistency with the given evidence -- and their refutation for the claim. 
% We thus designed to let human annotators evaluate these aspects by answering a series of questions. 
% We listed the questions presented to the labellers during the human evaluation stage and how we calculated those metrics in Appendix~\ref{appendix:annotation}. 
% In the real world, there is not one ground-truth counter-response for a given false claim, rather, there can be many -- each counter-response can counteract a different aspect of the claim -- and their semantics can be very different. 
% Therefore we focus on evaluating the factuality -- factual consistency with the given evidence -- and their refutation for the claim. Due to the cost limit of the evaluation, We evaluate a total of 100 instances for each dataset, ensuring a representative assessment of our approach. 
%
To evaluate the quality of counter-responses, we employed widely used LLM-based automatic metrics, which are scalable and have demonstrated human-comparable performance~\cite{liu2023g,min2023factscore}. 
%To assess the quality of the generated counter-responses.
We adapted G-EVAL~\cite{liu2023g} based on OpenAI's~\texttt{GPT-4o-mini} to evaluate four key dimensions: numerical accuracy, named entity accuracy, faithfulness, and refutation. G-EVAL is a framework for evaluating LLM outputs using metrics in a human-like evaluation criterion. We prompt \texttt{GPT-4o-mini} to produce scores on a scale of 1 to 5 for each of these dimensions,
%. For better comparison, we scale these scores to a 0-1 range for consistency across metrics. 
and scale the score to 0-1 to be consistent with other metrics. 
% The prompt used in G-EVAL can be found in Appendix~\ref{appendix:prompttemplate_geval}. 
% Recent research~\cite{falke2019ranking,zhou2021detecting} finds that traditional natural language generation evaluation metrics like ROUGE score~\cite{lin2004rouge} are poor at evaluating hallucinations in generated content. 
We also used FActScore~\cite{min2023factscore} to assess the factual precision. 
%as an automated factual precision evaluation metric. 
FActScore evaluates LLM outputs by breaking them down into atomic facts and verifying them against a knowledge source using LLMs. 
We use \texttt{GPT-4o-mini} as the base LLM, and the parameter $\gamma$ is set as 10 to penalize counter-responses with $<$10 atomic facts. 
We use the generated counter-responses from the test set to extract atomic facts and use the corresponding evidence as the knowledge source to score the atomic facts. 
We further compute an Overall score by averaging the scores from G-EVAL and FActScore. This comprehensive metric reflects both the factual accuracy and the effectiveness of the counter-responses in refuting the misinformation. 
% G-EVAL (numerical accuracy, named entity accuracy, faithfulness, and refutation)

% We also include a combined evaluation metric to reflect the performance by considering both the ability to refute the claim and the ability to generate factual content. This metric is named Overall and is done by averaging the value of Refutation and FActScore. 

%jz31: where are the human evaluation results? They are necessary. 
%In the real world, there is not one ground-truth counter-response for a given false claim, rather, there can be many -- each counter-response can counteract a different aspect of the claim -- and their semantics can be very different. 
In practice a false claim may have multiple valid counter-responses focusing on different aspects of the claim and varying in semantic content.
We therefore focus on evaluating the factuality -- factual consistency with the given evidence -- and their refutation for the claim. Due to the cost limit of evaluation, we evaluate a total of 100 claims for each dataset, ensuring a representative assessment of our approach.

\begin{table*}[t]
\begin{center}\small
\renewcommand{\arraystretch}{0.8}
\begin{tabular}{lcccccc}
\toprule
Model  & Numerical$\uparrow$ & Entity$\uparrow$ & Faithfulness$\uparrow$ & Refutation$\uparrow$ & FActScore$\uparrow$ & Overall$\uparrow$\\
\midrule
\multicolumn{6}{c}{PUBHEALTH Dataset} \\
\midrule
\multicolumn{5}{l}{$LLM_{gen}$ = Vicuna} \\ 
\hspace{4mm} MisMitiFact (ours)     & \textbf{0.987}  & 0.873   & \textbf{0.881}       & \textbf{0.716} & 0.733 & \textbf{0.838}\\
\hspace{4mm} MisMitiFact w/o NNE  & 0.926  & \textbf{0.879}          & 0.878       & 0.695 & \textbf{0.756} & 0.827 \\
\hspace{4mm} MisMitiFact w/o T  & 0.897 & 0.844 &  0.825 &  0.672  & 0.750 & 0.798 \\
\midrule
\multicolumn{5}{l}{$LLM_{gen}$ = LLaMA2} \\ 
\hspace{4mm} MisMitiFact (ours)    & 0.889    & 0.871          & 0.873       & 0.711  & 0.705 & 0.810 \\
\hspace{4mm} MisMitiFact w/o NNE  & 0.856    & \textbf{0.872}   & \textbf{0.890}       & \textbf{0.742}  & \textbf{0.708} & \textbf{0.813}\\
\hspace{4mm} MisMitiFact w/o T   & \textbf{0.922} &  0.842  &  0.835 &  0.714  & 0.626 & 0.788 \\
\midrule
\multicolumn{6}{c}{COVID-19 Vaccine Dataset} \\ 
\midrule
\multicolumn{5}{l}{$LLM_{gen}$ = Vicuna} \\ 
\hspace{4mm} MisMitiFact (ours)  & 0.987  & \textbf{0.911}   & \textbf{0.840}  & \textbf{0.690}  & 0.771 & \textbf{0.840}\\
\hspace{4mm} MisMitiFact w/o NNE     & \textbf{0.988}  & \textbf{0.911}          & 0.815       & \textbf{0.690} & \textbf{0.774} & 0.836\\
\hspace{4mm} MisMitiFact w/o T     &  0.980 & 0.872  & 0.775 & 0.683 & 0.770 & 0.816 \\
\midrule
\multicolumn{5}{l}{$LLM_{gen}$ = LLaMA2} \\ 
\hspace{4mm} MisMitiFact (ours)    & 0.933    & \textbf{0.869}   & \textbf{0.815}     & 0.717  & 0.686 & \textbf{0.804} \\
\hspace{4mm} MisMitiFact w/o NNE   & 0.955    & 0.866          & 0.796       & \textbf{0.721} &  0.674 & 0.802 \\
\hspace{4mm} MisMitiFact w/o T    &  \textbf{0.972}   &  0.855  &  0.773  & 0.692  & \textbf{0.707} & 0.800 \\
\bottomrule
\end{tabular}
\end{center}
\caption{The ablation study result for MisMitiFact}
\label{table:ablation}
\end{table*}

\subsection{Results}
% Our experiments are designed to test the effectiveness and the sensitivity of the proposed MisMitiFact to different data against several baselines. In this part, we show the experiment results on all the baselines and our proposed approach on both datasets. 

Table \ref{table:mainresult} shows the main experiment result. As can be seen, our MisMitiFact achieves robust performance, particularly in generating refutational counter-responses while maintaining faithfulness. Notably, MisMitiFact consistently matches or surpasses the best-performing baselines in numerical correctness, entity accuracy, faithfulness, and overall refutation quality. 
On PUBHEALTH, MisMitiFact achieves the highest average score (0.838) when using Vicuna as the $LLM_{gen}$, outperforming all baselines except for MisinfoCorrect w/ Evidence, which achieves a comparable score of 0.844. 
% Notably, MisMitiFact excels all other baselines, demonstrating its robustness to generate factually accurate and persuasive counter-responses. 
When using LLaMA2 as the $LLM_{gen}$, MisMitiFact maintains strong performance with an average score of 0.810, close to the best-performing baselines like SELF-REFINE (0.827) and MisinfoCorrect w/ Evidence (0.813). On COVID-19, MisMitiFact achieves the highest average score (0.840) with Vicuna as the $LLM_{gen}$, outperforming all baselines, including SELF-REFINE (0.828) and MisinfoCorrect w/ Evidence (0.830). 
%This underscores MisMitiFact's ability to handle noisy evidence and generate high-quality counter-responses. 
With LLaMA2 as the $LLM_{gen}$, MisMitiFact achieves an average score of 0.804, outperforming all baselines except SELF-REFINE, which achieves a comparable score of 0.806. While SELF-REFINE achieves competitive results in terms of FActScore, it is important to note that this approach leverages iterative refinement based on LLM feedback, which may introduce bias in favor of LLM evaluators. 
In summary, MisMitiFact demonstrates superior or comparable performance against the best-performing baselines across all datasets.

\begin{table*}[tbh]
\begin{center}\small
\begin{tabular}{lccccc}
\cline {1-5}
 & Numerical$\uparrow$ & Entity$\uparrow$ & Topic$\uparrow$  & Overall$\uparrow$ \\
\cline {1-5}
\multicolumn{5}{c}{PUBHEALTH} \\ 
\cline {1-5}
\scriptsize{$LLM_{gen}$=Vicuna} & 4.77 & 4.44 & 3.86 & 4.21  \\
\scriptsize{$LLM_{gen}$=LLaMA2} & 4.71 & 4.48 & 4.12 & 4.41 \\
\cline {1-5}
\multicolumn{5}{c}{COVID-19} \\ 
\cline {1-5}
\scriptsize{$LLM_{gen}$=Vicuna} & 4.98 & 4.33 & 2.93 & 3.36  \\
\scriptsize{$LLM_{gen}$=LLaMA2} & 4.96 & 4.38 & 4.08 & 4.18 \\
\cline {1-5}
\end{tabular}
\end{center}
%\caption{The LLM rating of critique model accuracy on the PUBHEALTH dataset and COVID-19 Vaccine dataset.}
\caption{Accuracy (5-point scale) of the critique models}
\label{table:critique_evaluate}
\end{table*}

\subsection{Accuracy and Efficiency of Critique Models}
%To further evaluate the effectiveness of MisMitiFacts, we will assess the accuracy of the critique models. 
%Due to the lack of ground truth data on critiquing the generated counter-response, we propose leveraging LLMs to conduct this evaluation. 
%By adopting a methodology akin to G-EVAL~\cite{liu2023g}, we utilize LLM ratings to evaluate the accuracy of the critique models. 
%It is not feasible to collect human annotations of textual critiques. 
% We adopted the methodology akin to G-EVAL~\cite{liu2023g} based on OpenAI's~\texttt{GPT-4o-mini} to rate how well the critique pinpointed the inconsistencies between the counter-response and the evidence on a scale of 1 to 5. 
We apply G-EVAL to evaluate how accurately the critique models pinpoint the inconsistencies between the counter-responses and evidence on a scale of 1 to 5. 
Table~\ref{table:critique_evaluate} shows the accuracy of the critique models for identifying errors in numbers, entities and topics. 
It can be seen that overall the critique model shows robust performance, especially when $LLM_{gen}$ = LLaMA2 (with overall ratings of 4.18-4.41). 
Critique models show the strongest performance on numerical errors (ratings 4.71-4.96 for Vicuna and LLaMA2 on both datasets), but varying performance for topic errors (from ratings of 2.93 for Vicuna on COVID-19 to 4.12 for LLaMA2 on PUBHEALTH). 
The informal social media language and diverse topics in the COVID-19 dataset may explain this. 

We assess the throughput of MisMitiFact critique models on a single machine with a single inference instance to evaluate their efficiency for large-scale applications. Throughput is measured in \#critiques/feedbacks per second, a metric that measures how many critique/feedback requests a model can process in one second. MisMitiFact’s critique models, based on T5-large (0.738B parameters), achieve a throughput of 0.925 critiques per second. In contrast, SELF-REFINE, which uses LLaMA2 (6.74B parameters) for feedback generation, achieves a throughput of only 0.165 critiques per second. 
%jz31: include the 5x speedup argument here. But 0.925/0.165 = 5.6 times.
MisMitiFact achieved a 5.6 times (0.925 vs. 0.165)  improvement in critique/feedback throughput, and this significant difference in throughput highlights the efficiency of MisMitiFact. 
MisMitiFact critique models are not only significantly smaller in size (0.738B vs. 6.74B parameters) but also faster in processing critiques, making them more suitable for large-scale misinformation mitigation. 

% speed compare - T5: 20.35 token/s LLaMA: 14.48 token/s
% lengh compare - T5: 22.01 token LLaMA: 87.56 token
% T5-large model size 0.738B  LLaMA2 model size 6.74B
% critics per second - T5: 0.925 /s LLaMA: 0.165 /s
%jz31: here needs to explain the lenght of the critiques (text length) in MisMitiFact vs. the feedback in REFINE to explain why. 
The high throughput of MisMitiFact can also be attributed to the simple, short critiques generated by MisMitiFact in contrast to the long narrative feedback generated by SELF-REFINE. 
As shown in the example in Fig.~\ref{fig:critique_example}, MisMitiFact generates critiques on numbers, facts and topics -- in 3 sentences of a total of 28 tokens. In contrast, SELF-REFINE generates a feedback narrative like ``Thank you for providing the claim and the facts. Here's my feedback on the original explanation: ... the explanation does not address the fact that males make up the majority of both the total count of people experiencing homelessness (57\%) and those sleeping rough (66\%)." -- in 5 sentences of a total of 124 tokens.

\subsection{Ablation Study}\label{sec:ablation}
%To evaluate the key components for better counter-response generation in MisMitiFact, 
We evaluated several variations of MisMitiFact with different parts of critique models: 
%jz31: rather than off-topic critieque model, change to topic-critique model to be sharp and succinct. 
\begin{itemize}
	\item \emph{MisMitiFact w/o number and named entity critique models (NNE)} is the variation that removes the number and named entity critique models and only includes the topic critique model.
	\item \emph{MisMitiFact w/o topic critique model (T)} is the variation that removes the topic critique model and only includes the number and named entity critique models. 
    %jz31: below argument is not strong.
    Number and named entity critique models are treated together since they both contribute to the faithfulness of the generated content.
\end{itemize}

Table \ref{table:ablation} shows the experiment result of the ablation study. We use the same evaluation metrics as in Table~\ref{table:mainresult}. It can be observed that MisMitiFact has the best performance on the overall performance in nearly all settings, the performance gain is achieved by both types of critiques. Removing NNE degrades numerical accuracy and faithfulness in specific settings (e.g., Vicuna on PUBHEALTH), while removing T degrades faithfulness (e.g., COVID-19 dataset). 
%jz31: below argument does not make sense to me. 
In FActScore, MisMitiFact has comparable performance to the best-performing settings, demonstrating the robust ability to generate counter-responses faithful to the evidence.

\section{Conclusion}
%jz31: i have rewritten to make it sharper and succinct. 
% This study addressed the problem of generating grounded responses to counter misinformation. 
% To generate factually accurate counter-responses, 
% we proposed a new paradigm for enhancing large language models through text-based critic-style feedback. 
% We proposed to employ simple, element-based critique feedback for LLMs to refine their generation of counter-responses and to ensure they are grounded in evidence. 
% The approach employs an ensemble of lightweight, to pinpoint errors in key elements such as numbers, named entities, and topics. 
% The training data is auto-generated from factual descriptions of related evidence collected from readily available fact-checking sites significantly reducing human effort. Experiment results demonstrate that this element-oriented, fine-grained feedback approach can generate comparable counter-responses to refute the misinformation against the best-performing baselines while using much smaller critique models and achieving a 5x increase in critique generation throughput, making our system highly suitable for large-scale misinformation counteraction. Our future work will focus on the generation of counter-responses in the presence of noisy evidence. 
% This study addressed the problem of generating grounded responses to counter misinformation. 
% To generate factually accurate counter-responses, 
% we proposed a new paradigm for enhancing large language models through text-based critic-style feedback.

This study tackled the challenge of generating grounded responses to counter misinformation. 
We proposed to generate simple critique feedback for LLMs to refine their initial generation and ensure responses are grounded in evidence. 
Our MisMitFact framework trains lightweight critique models on data sourced from readily available fact-checking sites to pinpoint errors in key elements like numbers, named entities, and topics in LLM generations. 
Experimental results show that MisMitiFact can generate counter-responses of comparable quality to LLMs' self-feedback while using significantly smaller critique models. 
Additionally, it achieves $\sim$5x increase in feedback generation throughput, making it highly suitable for cost-effective, large-scale misinformation mitigation. 
Our future work will focus on counter-response generation in the presence of noisy evidence. 

\cleardoublepage

\section*{Acknowledgments}

% Identification of funding sources and other support, and thanks to
% individuals and groups that assisted in the research and the
% preparation of the work should be included in an acknowledgment
% section, which is placed just before the reference section in your
% document.

% This section has a special environment:
% \begin{verbatim}
%   \begin{acks}
%   ...
%   \end{acks}
% \end{verbatim}
% so that the information contained therein can be more easily collected
% during the article metadata extraction phase, and to ensure
% consistency in the spelling of the section heading.

% Authors should not prepare this section as a numbered or unnumbered {\verb|\section|}; please use the ``{\verb|acks|}'' environment. 

%jz5: cannot make the below environment work
%\begin{verbatim}
%  \begin{acks}
This research is supported in part by the ARC Discovery Projects DP200101441, DP210100743 and ARC Linkage Project LP180100750.
%  \end{acks}
%\end{verbatim}

\bibliographystyle{named}
\bibliography{bibfile}

\end{document}